\documentclass[acmtog]{acmart}

\usepackage{booktabs} 
\usepackage{multirow} 
\usepackage{threeparttable}

\usepackage[ruled]{algorithm2e} 

\SetAlFnt{\small}
\SetAlCapFnt{\small}
\SetAlCapNameFnt{\small}
\SetAlCapHSkip{0pt}
\definecolor{lowSaturationGreen}{rgb}{0.5, 0.75, 0.5} 
\acmJournal{TOG}
\acmArticle{110}

\setcopyright{acmlicensed}
\copyrightyear{2026}
\acmYear{2026}
\acmVolume{45} 
\acmNumber{4}

\acmDOI{10.1145/3811306}

\def\ourname{ATGS}

\begin{document}
\title{\ourname: Anchored Temporal Gaussian Splatting for Long Volumetric Video Representation}

\author{Jiahao Wu$^*$}
\affiliation{%
 \institution{Guangdong Provincial Key Laboratory of Ultra High Definition Immersive Media Technology,
Shenzhen Graduate School, Peking University, Pengcheng Laboratory}
 \city{Shenzhen}
 \country{China}}
\email{2501111994@stu.pku.edu.cn}

\author{Jie Liang$^*$}
\affiliation{%
 \institution{Shenzhen Graduate School, Peking University, Pengcheng Laboratory}
 \city{Shenzhen}
 \country{China}}

\author{Die Hu}
\affiliation{%
 \institution{Shenzhen Graduate School, Peking University}
 \city{Shenzhen}
 \country{China}}

\author{Jiayu Yang}
\author{Xiaoyun Zheng}
\affiliation{%
 \institution{Pengcheng Laboratory}
 \city{Shenzhen}
 \country{China}}

\author{Kaiqiang Xiong}
\author{Xiang Li}
\affiliation{%
 \institution{Shenzhen Graduate School, Peking University, Pengcheng Laboratory}
 \city{Shenzhen}
 \country{China}}

\author{Chao Wang$^\dagger$}
\affiliation{%
 \institution{Pengcheng Laboratory}
 \city{Shenzhen}
 \country{China}}

 \author{Ronggang Wang$^\dagger$}
\affiliation{%
 \institution{Shenzhen Graduate School, Peking University, Pengcheng Laboratory}
 \city{Shenzhen}
 \country{China}}
\email{rgwang@pkusz.edu.cn}

\thanks{$^*$ equal contribution, $^\dagger$ corresponding author}

\begin{abstract}
Volumetric video enables immersive free viewpoint rendering of dynamic real world scenes, yet existing methods struggle with long sequences and complex motions, often leading to temporal instability and visual artifacts. To address these challenges, we propose \ourname, a Gaussian splatting based framework for volumetric video reconstruction. Our key insight is that explicitly tracking long term complex motion with individual Gaussian primitives is inherently unstable. Instead, we organize Gaussians around time conditioned anchors that localize their spatial and temporal support, thereby reducing long range motion complexity. We further introduce a temporal windowing strategy to activate only anchors relevant to the queried time, which improves scalability and temporal coherence. In addition, to ensure spatial and temporal stability, we design a compact set of multi level anchor features that encode global features, local spatial features, and local temporal features, jointly constraining Gaussian generation. Extensive experiments demonstrate that \ourname \ consistently outperforms prior methods on long sequence volumetric videos with complex motions. Project page: https://github.com/WuJH2001/ATGS. 
\end{abstract}

\begin{teaserfigure}
    \centering
    \includegraphics[width=\textwidth, trim=0cm 2cm 0cm 0cm]{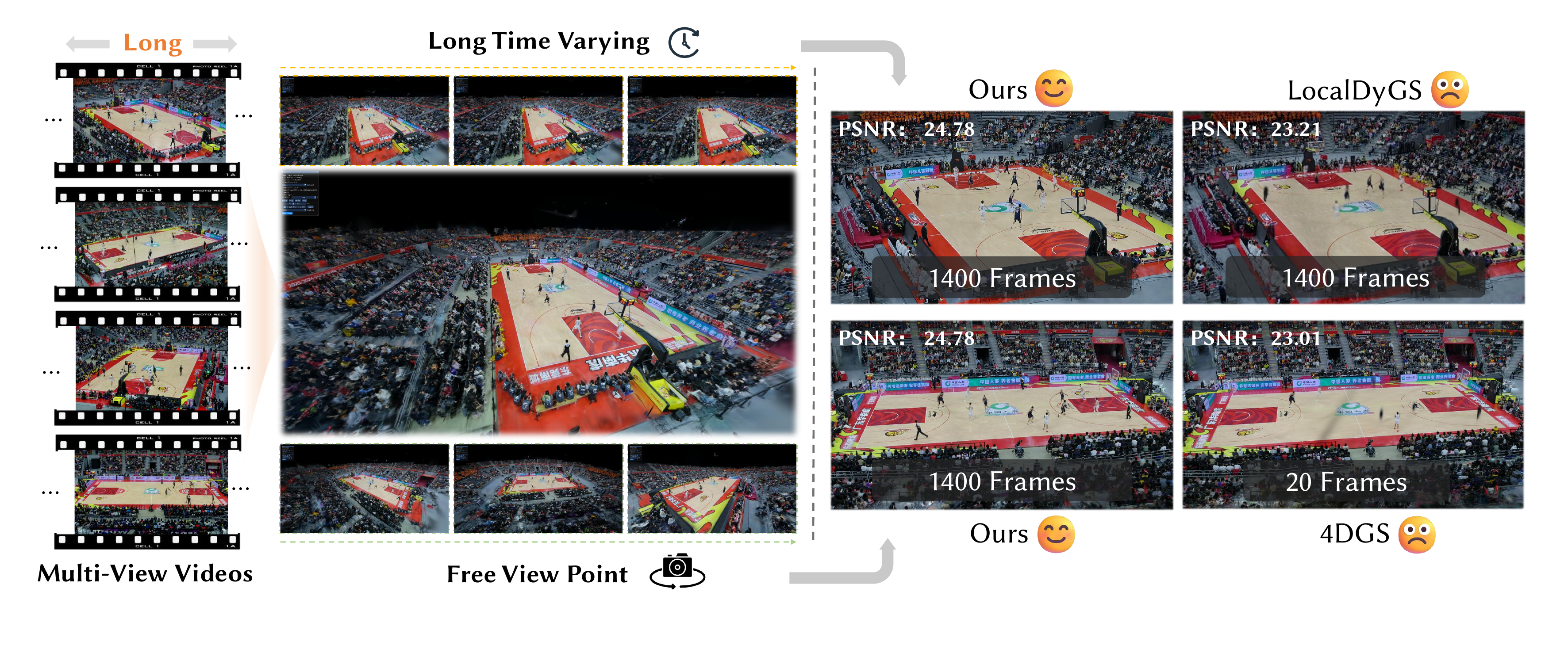}
    \caption{Based on long-sequence multi-view videos, our method is capable of reconstructing high-quality volumetric videos and enables novel view synthesis (NVS) at arbitrary viewpoints and time steps. Compared to prior approaches, which only support 20 frames fast motion or 300 frames minor hand or head motion (N3DV), our method supports fast motions over thousands.}
    \label{fig:teaser}
\end{teaserfigure}

%
%

%
%


\maketitle

\section{Introduction}
\label{sec: intro}

Volumetric video enables free viewpoint rendering of dynamic real world scenes, allowing users to observe events from arbitrary perspectives. This capability makes volumetric video a compelling medium for immersive visual experiences in a wide range of applications, including sports broadcasting, live events, gaming, VR, AR and XR \cite{rauschnabel2022xr,burdea2003virtual, speicher2019mixed,jin2023capture}.  Current volumetric video reconstruction pipelines typically convert multi-view 2D video inputs into 3D volumetric representations. However, they face two fundamental challenges: long-sequence reconstruction and complex motion modeling. Here, \textbf{long sequences} refer to videos with minute-level durations, while \textbf{complex motions} involve large inter-frame variations, such as fast multi-person movements from diverse viewpoints and dynamic lighting changes.  The goal of \ourname ~is to address both challenges in a unified framework.

With the rapid development of neural rendering \cite{mildenhall2021nerf,kerbl20233d}, data driven volumetric representations have become the dominant approach for volumetric video reconstruction. By jointly modeling geometry and appearance in a unified framework, neural rendering methods achieve high visual fidelity and flexible view synthesis without relying on explicit surface reconstruction or handcrafted priors. These advantages make neural rendering particularly suitable for reconstructing dynamic scenes from multi view captures. Existing neural rendering based approaches for volumetric video reconstruction can be broadly categorized into three classes. Deformation based approaches \cite{park2021hypernerf,park2021nerfies,wuswift4d,wu20244d,yang2024deformable,huang2024sc} model dynamic scenes by learning deformation fields that warp a canonical space over time. 4D Gaussian based methods \cite{yang2023real,xu2024longvolcap,wang2025freetimegs,4drotorgs} represent dynamics using explicit spatio temporal Gaussian primitives, enabling efficient rendering with high visual fidelity. 4D grid based methods \cite{cao2023hexplane,xu20234k4d,kplanes} adopt compact spatio temporal grids to store implicit features, which are decoded to recover geometry and appearance at each sample. Despite their success on short sequences and controlled settings, these approaches remain challenged when scaling to large scale dynamic scenes with long temporal durations and complex motions, where temporal instability, motion drift, and visual artifacts often emerge.

These limitations have also been recognized by several recent methods. LongVolCap \cite{xu2024longvolcap} focuses on long sequence volumetric video reconstruction and demonstrates the ability to model minute level captures, but its performance degrades under complex and fast changing motions \cite{wang2025freetimegs}. In contrast, methods such as FreeTimeGS \cite{wang2025freetimegs} and LocalDyGS \cite{wu2025localdygs} are designed to handle complex dynamics, yet are limited to short temporal windows of only one to two seconds. Simply splitting long sequences into independent short clips leads to temporal inconsistencies and flickering artifacts. Consequently, jointly modeling long temporal durations and complex motions remains an open and challenging problem in volumetric video reconstruction.

To address these challenges, we propose \ourname, a volumetric video reconstruction method that jointly targets long temporal durations and complex motions. A fundamental difficulty in this setting lies in explicitly modeling complex motion over long sequences, which often leads to unstable optimization and accumulated errors. Rather than directly tracking long range motion, our method introduces a set of time conditioned anchors that constrain the spatial and temporal Gaussian generation, effectively decomposing long term dynamics into a collection of temporally localized modeling tasks.
To further control temporal complexity, we employ a temporal windowing strategy that activates only anchors relevant to the queried time during training and inference. This design limits interference from distant time steps and improves temporal coherence, enabling stable optimization over long sequences.

To further stabilize Gaussian generation, we introduce a hierarchical anchor feature formulation that progressively constrains spatial structure and temporal variation.
We start by associating each time conditioned anchor with a global spatial feature, which provides coarse contextual information about its surrounding region. While this representation offers strong expressive power, it alone does not enforce local spatial consistency, making optimization susceptible to temporal jitter over long sequences.
To address this limitation, we augment the anchor with local spatial features that encode locally consistent, time invariant structure. These features complement the global representation by constraining Gaussian generation within a stable spatial neighborhood, thereby improving robustness during training.
Finally, to capture dynamic variations without introducing long range temporal dependencies, we incorporate local temporal features. By conditioning these features on the anchor position and local time within short temporal intervals, the model represents motion in a temporally localized manner.
Together, the global spatial, local spatial, and local temporal features form a unified anchor representation that is jointly decoded to generate temporally aware Gaussian primitives around each anchor, enabling photorealistic free viewpoint rendering.

We evaluate our method on a diverse set of long-sequence, multi-view datasets, including N3DV (Flame Salmon) \cite{li2022neural}, VRU \cite{wuswift4d}, and MeetRoom. In addition, we validate the generalization capability of our approach on the SelfCap dataset \cite{xu2024longvolcap}, which contains several thousand frames of video. Please refer to the demonstration video and the supplementary material for additional results.

Our contributions can be summarized as:

\noindent (1) We propose a temporal anchor based Gaussian representation for volumetric video reconstruction, which localizes Gaussian primitives in space and time through time conditioned anchors and temporal windowing. This representation enables effective modeling of long sequence dynamic scenes with complex motions.

\noindent (2) We introduce a hierarchical anchor feature formulation that progressively constrains Gaussian generation using global spatial, local spatial, and temporal features. This design improves spatial stability and temporal consistency, leading to higher quality reconstruction of dynamic scenes.

\noindent (3) We conduct extensive experiments on multiple challenging benchmarks, demonstrating that our method consistently outperforms existing approaches on volumetric video reconstruction involving long temporal durations and complex motions.

\section{Related Work}
\label{sec:related work}

\subsection{Volumetric video}

Volumetric video is a 3D video representation that captures depth and color information of dynamic scenes to produce a sequence of three-dimensional models viewable from arbitrary viewpoints. By leveraging multi-view camera setups, it overcomes the planar limitations of conventional 2D video and enables full 6-DoF free-viewpoint rendering. Early approaches relied heavily on strong priors, such as shape-from-silhouette techniques \cite{ahmed2008dense}, deformable template models \cite{carranza2003free}, or template-free dynamic reconstruction systems augmented with depth sensors \cite{newcombe2015dynamicfusion}. There also exist generative approaches \cite{lombardi2019neural,bozic2020deepdeform} aimed at improving the quality of dynamic reconstruction; however, these methods either struggle with real-world scenes that involve complex structures and severe occlusions, or rely on expensive and sophisticated capture systems. Together, these limitations hinder the practical deployment of volumetric video technologies in real-world applications.

\subsection{Neural radiance field for dynamic scenes}

In recent years, Neural Radiance Fields (NeRF)\cite{mildenhall2021nerf} have attracted significant attention due to their ability to deliver photorealistic and immersive visual experiences. Subsequent works \cite{park2021hypernerf, park2021nerfies, pumarola2021d}  have extended NeRF to dynamic scenarios and volumetric video reconstruction. For example, D-NeRF \cite{pumarola2021d} and NeRFies \cite{park2021nerfies} adopt deformation fields to warp a canonical space over time, enabling dynamic view synthesis, while \cite{du2021neural} explicitly incorporates the temporal dimension into the spatial domain to model spatiotemporal variations. 
However, the high training cost of NeRF-based representations motivates the development of more efficient alternatives. Grid-based representations that enable faster reconstruction have therefore gained popularity, such as Plenoxel and DVGO \cite{ fridovich2022plenoxels, dvgo}. Building upon these representations, several dynamic extensions have been proposed, including MixVoxel \cite{wang2023mixed}, which further accelerates reconstruction. Despite their efficiency, these methods suffer from high memory consumption, which limits their scalability and practical deployment.
To address this issue, Instant-NGP  introduces a multi-resolution hash encoding to store node features efficiently, significantly reducing memory usage. This idea has been further extended to dynamic settings, for example in MSTH \cite{wang2023masked}. In parallel, researchers have explored factorizing 3D or 4D volumetric representations into lower-dimensional structures. Notably, Tensorf \cite{chen2022tensorf} decomposes volumetric fields into compact tensor representations, while explicit multi-plane methods such as K-Planes, HexPlane, and 4K4D \cite{cao2023hexplane,kplanes,xu20244k4d} represent 4D scenes using a set of structured planes, achieving a favorable balance between efficiency and expressiveness.

\subsection{Gaussian splatting for dynamic scenes}

3D Gaussian Splatting (3DGS) \cite{kerbl20233d} has attracted significant attention due to its ability to efficiently reconstruct high-fidelity 3D scenes from posed images. A series of extensions have been proposed to further improve surface reconstruction quality \cite{zhang2024rade,huang20242d}, accelerate training \cite{taminggs}, and improve rendering quality \cite{lu2024scaffold, yu2024mip}.
In the dynamic setting, several representative research directions have also emerged, including frame-wise training methods \cite{sun20243dgstream,luiten2024dynamic}, deformation-based methods \cite{wu20244d,yang2024deformable,huang2024sc}, and 4D Gaussian–based approaches \cite{yang2023real,duan20244d,wang2025freetimegs,xu2024longvolcap}. Frame-wise training methods can handle long video sequences, but they require substantial storage and often suffer from inter-frame error accumulation, which may lead to unstable training or model collapse in later stages. Deformation-based methods, on the other hand, struggle to model complex motions and to robustly track long-term dynamics.

More recently, several 4D Gaussian–based methods \cite{wang2025freetimegs,xu2024longvolcap,liang2026clipgstream} have attempted to address either long-sequence reconstruction or complex motion modeling. For example, LongVolCap \cite{xu2024longvolcap} decomposes long-term motion into multiple hierarchical levels using layered 4D Gaussians, but shows limited capability in handling highly complex motions. In contrast, FreeTimeGS \cite{wang2025freetimegs} places a large number of 4D Gaussian primitives in space to better capture complex dynamics, while LocalDyGS \cite{wu2025localdygs} adopts an anchor-based formulation to improve motion modeling. However, these methods are restricted to handling only second-level complex motions and fail to scale to minute-level long-sequence volumetric video reconstruction. 
To address these limitations, we propose \ourname, which is designed to jointly tackle the challenges of complex motion modeling and long-duration volumetric video reconstruction.

\begin{figure*}[t]
    \centering
    \includegraphics[width= \textwidth ,trim=0.5cm 0.5cm 0.5cm 0cm]{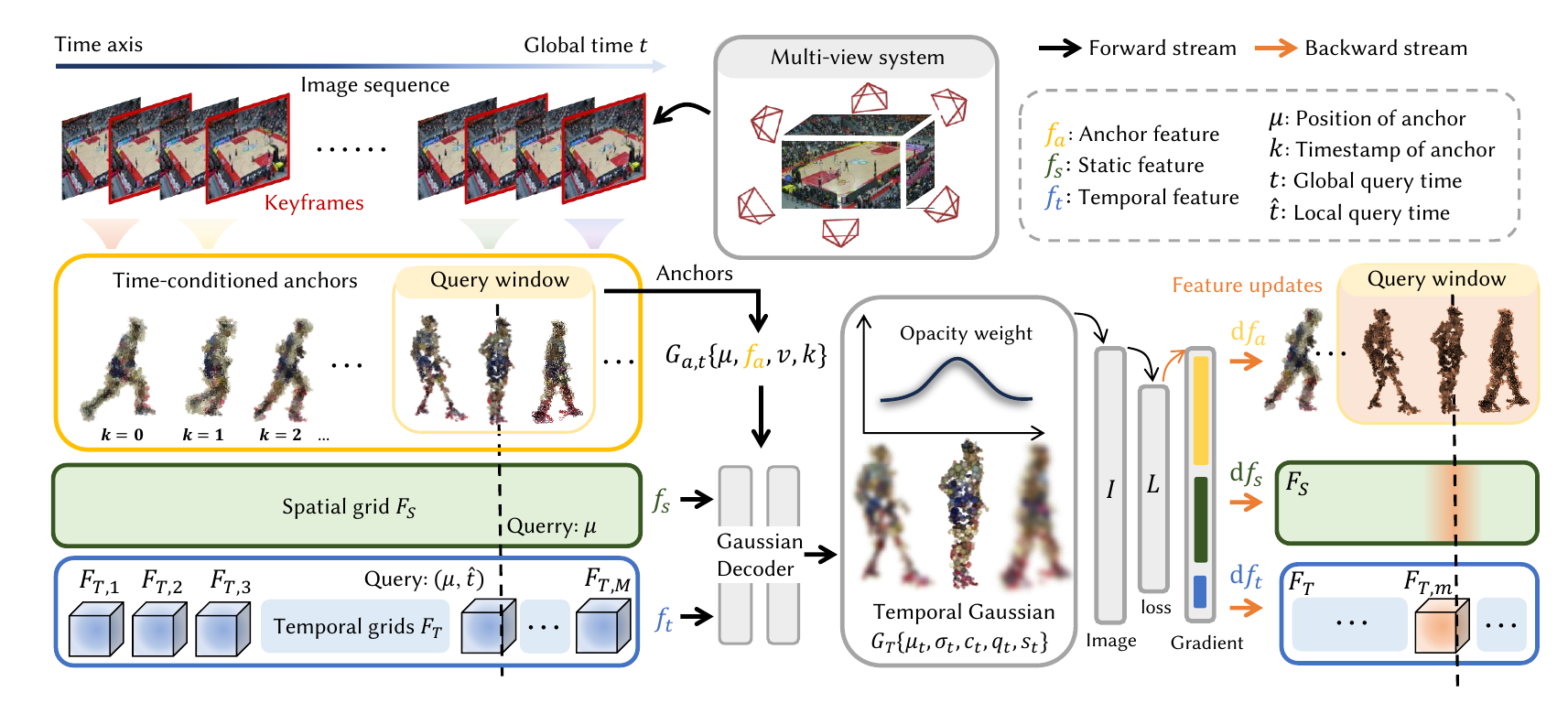}
    \vspace{-2em}
    \caption{\textbf{ Overview of  \ourname.} Keyframes are extracted from each view to initialize time-conditioned anchors with independent anchor features. Anchors query spatial and temporal grids using their positions and timestamps, and the fused features are decoded into temporal Gaussians. During training, only anchors and temporal grids associated with the current timestamp are updated.}
    
    \vspace{-1.5em}
    \label{figs:pipelins}
\end{figure*}

\section{Method}
\label{sec: method}

\subsection{Overview}

Our method targets volumetric video reconstruction with long temporal durations and complex motions by representing dynamic scenes using temporally localized Gaussian primitives. The framework consists of four main components.
We first introduce time-conditioned anchors to represent the dynamic scene over long sequences (Sec.~\ref{subsec: anchor initial}). A temporal windowing strategy is then employed to select anchors relevant to a queried time, ensuring smooth temporal evolution during inference (Sec.~\ref{subsec: temporal window}). To enable stable and expressive modeling, each anchor is augmented with a hierarchical spatio-temporal feature representation (Sec.~\ref{subsec: anchor feature}). Finally, the enriched anchors are decoded into Temporal Gaussians for rendering (Secs.~\ref{subsec: temporal gaussian}), and the entire model is trained end-to-end with image-based losses and regularization (Secs.~\ref{subsec: loss}).

\subsection{Time-conditioned Anchor Initialization}
\label{subsec: anchor initial}

To model long sequence, large scale spatiotemporal scenes, we represent the dynamic content using a sparse set of anchors initialized from periodically sampled keyframes. This strategy provides sparse yet comprehensive coverage of scene geometry over time, ensuring that dynamic objects throughout the sequence are associated with anchor representations.

Each anchor is defined by a set of spatial and temporal attributes. Specifically, we assign an anchor position \( \mu \in \mathbb{R}^3 \) to represent its 3D location in space, together with a learnable anchor feature \( f_a \in \mathbb{R}^{64} \) that encodes local spatial characteristics around the anchor. In addition, each anchor is associated with a spatial scale parameter \( v \in \mathbb{R}^3 \), which defines the spatial extent of its influence and serves as a normalization factor for subsequent Gaussian generation. To incorporate temporal information, we further assign each anchor a discrete temporal index \( k \in \{1,\ldots,K\} \), indicating the keyframe from which it is initialized.
Anchors equipped with both spatial and temporal attributes are referred to as \emph{time-conditioned anchors}. Formally, a time-conditioned anchor is defined as
\begin{equation}
G_a = \{\mu, f_a, v, k\}.
\end{equation}

\textbf{This time-conditioned anchor formulation significantly simplifies the modeling of complex motions.} By using anchors as priors, the model no longer needs to explicitly model Gaussian point trajectories; instead, anchors control the appearance and disappearance of Gaussian primitives, which is sufficient to achieve perceptually plausible motion modeling.

\subsection{Temporal Windowing Strategy}
\label{subsec: temporal window}

As shown in Fig.\ref{figs:pipelins}, given the set of time-conditioned anchors defined over the entire sequence, we introduce a \emph{temporal windowing strategy} to select anchors relevant to a queried time \( t \in [0,1) \) while ensuring temporally coherent inference.

We define a temporal window of size \( W \in \mathbb{N} \) (default $W=7$) and denote its half-width as
\begin{equation}
h = \left\lfloor \frac{W}{2} \right\rfloor .
\end{equation}
For a given query time \( t \), we identify the nearest keyframe index \( \bar{k} \) by mapping the normalized time to the keyframe index domain,
\begin{equation}
\bar{k} = \arg\min_{j \in \{1,\ldots,K\}} \left| t \cdot (K-1) + 1 - j \right|.
\end{equation}
The anchor group associated with time \( t \) is then constructed by collecting anchors whose temporal indices fall within a window centered at \( \bar{k} \),
\begin{equation}
G_{a,t} =
\left\{
G_a\{\mu, f_a, v, k\}
\;\middle|\;
k \in \left[ \max(1,\bar{k}-h),\ \min(K,\bar{k}+h) \right]
\right\}.
\end{equation}

This temporal windowing strategy ensures that the anchor set used for inference evolves smoothly over time, avoiding abrupt changes in anchor activation across neighboring frames. As a result, it stabilizes long-sequence rendering and effectively reduces temporal jitter, which is a common challenge in volumetric video reconstruction over extended temporal durations.



\subsection{Spatial-Temporal Feature Representation}
\label{subsec: anchor feature}

\begin{figure*}[t]
    \centering
    \includegraphics[width= \textwidth ,trim=0cm 1cm 0cm 0cm]{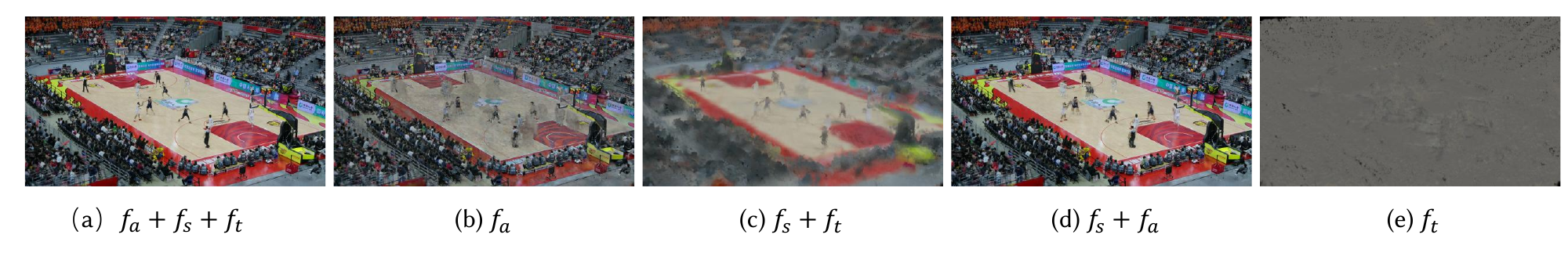}
    \vspace{-2em}
    \caption{Visualization of the information captured by each feature component after training. As shown in (a), the combination of all features enables full image rendering. (b,c) indicate that anchor features $f_a$ capture most of the scene information, while the other components ($f_s,f_t$) contribute marginally. (d,e) show that anchor and static features together encode nearly all scene content (all Gaussians), while dynamic features primarily control the appearance and disappearance of Gaussian primitives rather than explicitly modeling the scene.}
    \vspace{-1.2em}
    \label{figs: features}
\end{figure*}

Given the anchor group selected at a queried time \( t \),
\begin{equation}
G_{a,t} =
\left\{
G_a\{\mu, f_a, v, k\}
\;\middle|\;
k \in \left[\max(1,\bar{k}-h),\ \min(K,\bar{k}+h)\right]
\right\},
\end{equation}
we next enrich each anchor with additional spatial and temporal features. The goal is to enable expressive temporal modeling while enforcing local spatial consistency to stabilize training over long sequences.

\paragraph{Local Consistency Representation.}
Our anchor representation captures local scene information using discrete anchor features, with each anchor optimized independently. While this design is sufficient for static reconstruction or short sequences, it often leads to training instability and motion blur when applied to long volumetric videos. To mitigate this issue, we introduce a shared spatial grid \( F_s \) to enforce local spatial consistency across anchors.

Specifically, each anchor queries the spatial grid via trilinear interpolation at its spatial location \( \mu \) to obtain a static spatial feature
\begin{equation}
f_s = F_s(\mu), \quad F_s \in \mathbb{R}^{X \times Y \times Z \times D},
\end{equation}
where \( (X, Y, Z) \) denote the spatial resolution of the grid and \( D \) is the feature dimension at each grid node. These spatial features provide locally consistent and time-invariant cues that complement the anchor features \( f_a \), improving stability during optimization.

\paragraph{Temporal Dynamics Modeling.}
Modeling temporal variation in volumetric video has been approached using diverse representations, including MLP-based formulations, multi-plane factorizations, and hash-grid structures. \textbf{However, using a single temporal representation is often insufficient to capture long-range dynamics in extended video sequences.}

To address this limitation, we partition the temporal domain into \( M \) (we set $M=12$ in VRU dataset, others $M=2$, details in appendix) segments and introduce a set of temporal structures
\[
F_T = \{F_{T,1}, \ldots, F_{T,M}\},
\]
where each structure \( F_{T,m} \) models dynamic variations within the temporal interval \( [(m-1)/M, m/M] \). In our implementation, each \( F_{T,m} \) is instantiated as a 4D hash grid.

Given a query time \( t \in [0,1) \), the corresponding temporal segment index is computed as
\begin{equation}
m = \left\lfloor M \cdot t \right\rfloor + 1,
\end{equation}
and the local time coordinate within the segment is
\begin{equation}
\hat{t} = M \cdot t - m + 1.
\end{equation}
Each anchor then queries the associated temporal structure via quadrilinear interpolation to obtain a temporal feature
\begin{equation}
f_t = F_{T,m}(\mu, \hat{t}).
\end{equation}
This formulation decouples local temporal modeling from the global sequence length, enabling efficient and stable representation of long-term dynamics.

\paragraph{Augmented Anchor Representation.}
Finally, each anchor in the group \( G_{a,t} \) is augmented with both spatial and temporal features, yielding the enriched anchor representation
\begin{equation}
G_{a,t} =
\left\{
G_a\{\mu, f_a, f_s, f_t, v, k\}
\;\middle|\;
k \in \left[\max(1,\bar{k}-h),\ \min(K,\bar{k}+h)\right]
\right\}.
\end{equation}
These augmented anchors serve as the input to the subsequent temporal Gaussian decoding stage.

The roles of the three feature components are illustrated in Fig.~\ref{figs: features}. The anchor and static features are responsible for encoding nearly all scene content, while the dynamic features capture only a small amount of time-varying information and do not explicitly contribute to the representation of static scene geometry. 
As details below:

(1) Anchor feature $f_a$. Following Scaffold-GS \cite{lu2024scaffold}, anchors are decoded into Gaussians: $G=Decoder(f_a)$. However, time-invarying anchor features alone cannot model dynamic scenes.

(2) Temporal feature $f_t$. Therefore, we introduce a shared hash grid $F_T(x,y,z,t)$ to provide time-aware features, enabling time-varying Gaussians for dynamic scene reconstruction:$G_t=F_*(f_a,f_t)$.

\begin{figure*}[h]
    \centering
    \includegraphics[width= \textwidth ,trim=0cm 1cm 0cm 0cm]{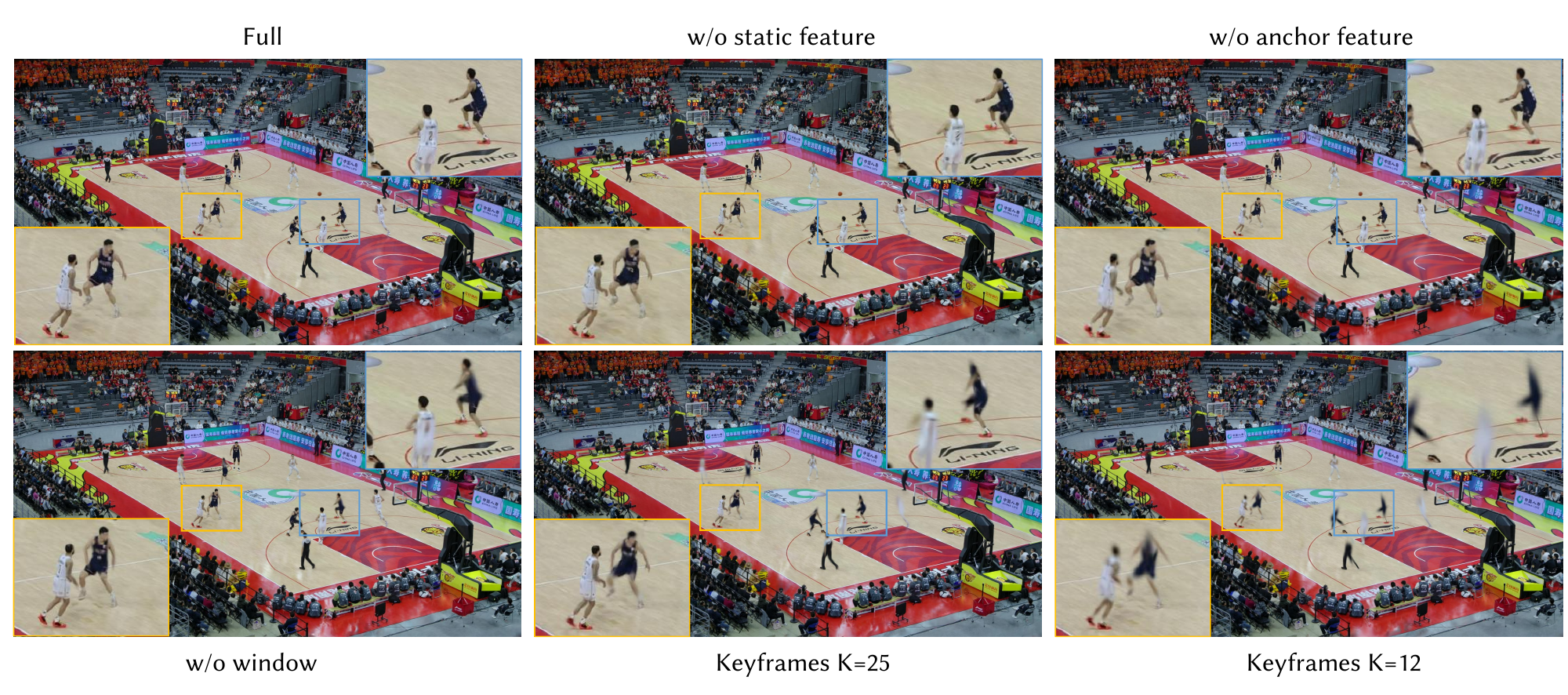}
    \caption{Additional qualitative visualizations from the ablation study (On VRU Long 250 frames).}
    \label{figs: ablation}
\end{figure*}

(3) Static feature $f_s$. Since $f_a$ is optimized independently across anchors without spatial locality, it can lead to blurry reconstructions in complex scenes(Fig. \ref{figs: ablation}). We therefore introduce $F_S(x,y,z)$ to encode low-frequency spatial structure and improve stability.

In summary, $F_T$ models temporal variation while $F_S$ provides stable spatial structure. Ablation on the VRU dataset (Tab. \ref{tab: abla features}) shows that removing any component degrades performance.

\subsection{Temporal Gaussian Decoding}
\label{subsec: temporal gaussian}

Given a query time \( t \), the proposed pipeline yields the corresponding anchor group \( G_{a,t} \) together with enriched spatio-temporal features. We decode these features using a Gaussian decoder \( F_* \) to generate a set of \emph{Temporal Gaussians}, whose parameters vary continuously with time:
\begin{equation}
G_T\{\mu_t, q_t, s_t, \sigma_t, c_t\} = F_*(G_{a,t}).
\end{equation}

Each time-conditioned anchor generates \( q \) Temporal Gaussians. The mean positions of these Gaussians are computed as
\begin{equation}
\{\mu_t^i\}_{i=0}^{q-1} = \mu + v \cdot F_{\mu}(f_w),
\quad
f_w = [\, f_a + f_s \,;\, f_t \,],
\end{equation}
where \( \mu \) and \( v \) denote the anchor position and spatial scale, respectively, as defined in Sec.~\ref{subsec: anchor initial}. The function \( F_{\mu}(\cdot) \) is a lightweight MLP that outputs a vector of size \( q \times 3 \), following prior Gaussian-based formulations~\cite{wu2025localdygs,lu2024scaffold}. Here, \( \mu_t^i \) represents the mean of the \( i \)-th Temporal Gaussian generated by the anchor at time \( t \). 
We adopt $f_w = [\, f_a + f_s \,;\, f_t \,]$ instead of $f_w = f_a + f_s + f_t$, since the latter requires implicit disentanglement of static and dynamic components, which incurs an approximate 10\% training slowdown. In contrast, the adopted formulation facilitates decoupled optimization of static and temporal features, leading to improved training efficiency and stability.

All remaining Gaussian attributes, including orientation \( q_t \), scale \( s_t \), opacity \( \sigma_t \), and color \( c_t \), are decoded in an analogous manner from the same spatio-temporal feature representation. The resulting set of Temporal Gaussians is then used for volumetric rendering.

\subsection{Loss Function}
\label{subsec: loss}

We train the proposed model using a combination of image-based reconstruction losses and a compactness regularization term, following common practice in Gaussian-based rendering methods~\cite{kerbl20233d,lu2024scaffold,wu2025localdygs}.
To encourage each Temporal Gaussian to remain spatially localized around its associated anchor, we impose a volume-based regularization on the Gaussian scales:
\begin{equation}
\mathcal{L}_v = \sum_{i=1}^{P} \mathrm{Prod}(s_t^i),
\end{equation}
where \( P \) denotes the number of active Temporal Gaussians across all anchors at query time \( t \), \( s_t^i \in \mathbb{R}^3 \) represents the scale of the \( i \)-th Temporal Gaussian, and \( \mathrm{Prod}(\cdot) \) computes the product of the scaling components. This regularization penalizes overly large Gaussians and promotes compact, anchor-centered representations.
In addition to the compactness constraint, we adopt standard image reconstruction losses, including the \( L_1 \) loss and the structural similarity loss \( \mathcal{L}_{\text{SSIM}} \). The overall training objective is defined as
\begin{equation}
\mathcal{L}
= (1 - \lambda_{\text{SSIM}})\,\mathcal{L}_1
+ \lambda_{\text{SSIM}}\,\mathcal{L}_{\text{SSIM}}
+ \lambda_v\,\mathcal{L}_v,
\end{equation}
where \( \lambda_{\text{SSIM}} = 0.2 \) and \( \lambda_v = 0.001 \) balance the contributions of the reconstruction and regularization terms.

\section{Experiments}
\label{sec: experiment}

\begin{figure*}[h!]
    \centering
    \includegraphics[width= \textwidth ,trim=0cm 0cm 0cm 0cm]{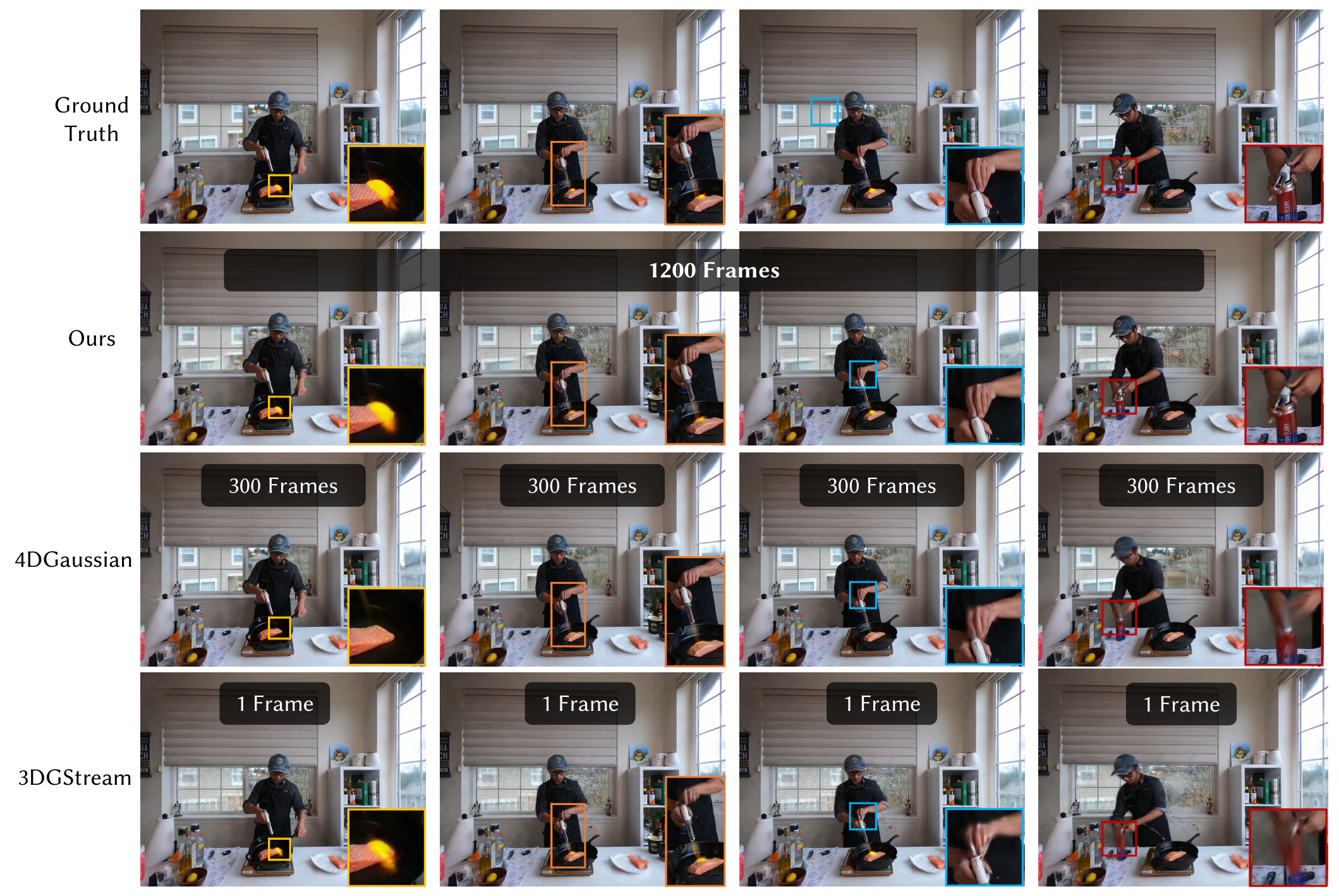}
    \vspace{-2.5em}
    \caption{Qualitative comparisons on the N3DV dataset. Our model can reconstruct a 40-second video (1,200 frames) in a single run and achieves higher rendering quality than other methods operating on much shorter segments.}
    \label{figs:comparison_n3dv}
\end{figure*}

\subsection{Implementation}

For our method, the dimensions of features $f_a$, $f_s$, and $f_t$ are set to 64. 
The resolution and feature dimensionality of $F_S$ are set to $\{X, Y, Z, D\} = \{128, 128, 128, 64\}$. 
The hash table size of each $F_T$ is set to \( 2^{16} \) to trade off performance against storage, with other settings consistent with INGP \cite{muller2022instant}. All MLPs in the Gaussian Decoder consist of two layers with ReLU activation, followed by Sigmoid or normalization at the output. 

Our method uses the Adam optimizer~\cite{kingma2014adam}, similarly to 3DGS. 
We employ a fixed learning rate of 0.0075 for the features $f_a$ and $f_s$. 
An independent learning rate decay strategy is applied to each temporal grid $F_t$, ensuring that only the currently accessed grid undergoes decay.

In addition, we have two important parameters: the number of keyframes \(K\) and the number of temporal grids \(M\), which are adjusted according to the magnitude of scene motion. For scenes with small motion, such as N3DV and MeetRoom, we set \(K=15\) and \(M=2\); only a few keyframes and temporal grids are sufficient. In contrast, for scenes with large motion, such as VRU, which contain a large amount of dynamic information, a small number of temporal grids cannot capture the extensive temporal variations. Therefore, we set \(K=250\) and \(M=12\) for such cases. The temporal window size \(W\) is fixed to 7 for all scenes. All experiments are conducted on NVIDIA A100 GPU(80~GB). 

\subsection{Datasets}

To comprehensively evaluate our method, we conduct experiments on several publicly available datasets with diverse temporal lengths and motion complexities, including the N3DV, VRU, and MeetRoom. Notably, the latter two datasets pose significantly greater challenges due to their temporal scale and motion variation.

\textbf{N3DV \cite{li2022neural}}. Neural3DV is a widely used benchmark for novel view synthesis, captured using a multi-view system with 19–21 cameras. The videos are recorded at a resolution of \(2704\times2028\) at 30 FPS. The dataset includes five scenes with 300 frames and one scene with 1200 frames. Following previous work, we downsample the data and split the cameras into training and testing sets.

\begin{table}[t]
    \centering
        \caption{\textbf{Quantitative comparison on the MeetRoom dataset \cite{li2022streaming}.} PSNR is averaged across all 300 frames, while training time and storage requirements accumulate over the entire sequence. }
        \vspace{-0.8em}
    \begin{threeparttable}
    \resizebox{0.8\columnwidth}{!}
    {
        \begin{tabular}{lcccr@{}}
        \toprule
         Method & PSNR \(\uparrow\)  & Time(hours) \(\downarrow\)  & Size(MB) \(\downarrow\) \\
        \midrule[0.5pt]
        \multirow{1}{*}{StreamRF \cite{li2022streaming}}  & 26.72 & 0.85 & 2700 \\
        \multirow{1}{*}{4DGaussian \cite{wu20244d}}  & 30.27 & 0.50 & \textbf{80} \\
        \multirow{1}{*}{3DGStream \cite{sun20243dgstream}}  & 30.79& 0.60 & 1230 \\
        \multirow{1}{*}{Hicom \cite{hicom2024}}  & 26.73 & \textbf{0.20}  & 180 \\
        \multirow{1}{*}{\textbf{\ourname(Ours)}}  & \textbf{32.79} & 0.45 & 110  \\
        \bottomrule
        \end{tabular}
    }
\end{threeparttable}
\vspace{-1em}
    \label{tab: meetroom}
\end{table}

\begin{table}
\centering

\caption{
    \textbf{Quantitative comparisons on the Neural 3D Video Dataset \cite{li2021neural}.}
    ``Size'' is the total model size for 300 frames. DSSIM$_1$ sets data range to 1.0 while DSSIM$_2$ to 2.0 \cite{li2024spacetime}. $\ast$ indicates online method. Tested on 300 frames.
}
\vspace{-1.2em}
\label{tab: n3dv}
\resizebox{\linewidth}{!}{
\begin{tabular}{@{}lccccccccc}
    \toprule
   Method & PSNR$\uparrow$ & DSSIM$_1$$\downarrow$& DSSIM$_2$$\downarrow$ & LPIPS$\downarrow$ & FPS$\uparrow$ & Time$\downarrow$ \\
  \midrule
  StreamRF~$^\ast$~\cite{li2022streaming}  & 28.26 & - & - & - & 10.9 & -\\
  NeRFPlayer~\cite{song2023nerfplayer}  & 30.69 & 0.034 & - & 0.111 & 0.05 & 6.0 h  \\
  HyperReel~\cite{attal2023hyperreel}  & 31.10 & 0.036 & - & 0.096 & 2 & -\\
  K-Planes~\cite{kplanes}    &  31.63 &  0.018  & -  & - & 0.3  & 5.0 h \\
  HexPlane~\cite{cao2023hexplane}    &  31.70 &  0.014   & - & 0.075 & 0.21  & 12.0 h \\
  MixVoxels~\cite{wang2022mixed}   & 31.73  & - &  0.015   & 0.064 & 4.6 & - \\
  4DGaussian~\cite{wu20244d}     & 31.02 &0.030  & - & 0.150  & 30  & 0.67 h \\
  3DGStream$\ast$~\cite{sun20243dgstream}     & 31.67 & -  & - & -  & \textbf{215}  & 1.0 h \\
 RealTimeGS~\cite{yang2023real}     &  32.01 & -  & 0.014 & 0.055  & 114  & 9.0 h   \\
  SpaceTimeGS ~\cite{li2024spacetime}         & 32.05 & 0.026  & 0.014 &0.044  & 140 & $>5$ h \\
  LocalDyGS ~\cite{wu2025localdygs}         & 32.28 & 0.028  & 0.014 &0.044  & 105 & \textbf{0.58 h} \\
   \textbf{\ourname(Ours)}         & \textbf{32.56} & \textbf{0.027}  & \textbf{0.013} & \textbf{0.043}  & 70 & 0.9 h  \\
  \bottomrule
\end{tabular}
}
\vspace{-1.5em}
\end{table}


\begin{table}[htbp]
\centering
\caption{
\textbf{Quantitative comparison on \textit{VRU-basketball}~\cite{wuswift4d}.}
We report PSNR, and SSIM to evaluate rendering quality.
Methods marked with $^1$ are trained on short segments (20 frames or single-frame),
while methods marked with $^2$ are trained on long segments (250 frames or 1400 frames).
The best results are highlighted in bold.
}
\vspace{-0.6em}
\label{tab:vru}
\resizebox{\linewidth}{!}{
\begin{tabular}{l|cc|cc|cc}
\toprule
\multirow{2}{*}{Methods} 
& \multicolumn{2}{c|}{VRU GZ} 
& \multicolumn{2}{c|}{VRU DG} 
& \multicolumn{2}{c}{VRU Long} 
\\
\cmidrule(lr){2-3} \cmidrule(lr){4-5} \cmidrule(lr){6-7}
& PSNR $\uparrow$ & SSIM $\uparrow$ & PSNR $\uparrow$ & SSIM $\uparrow$ 
& PSNR $\uparrow$ & SSIM $\uparrow$ \\
\midrule
3DGS$^1$        & 30.50  & \textbf{0.949}  & 30.28 & \textbf{0.945}  & - & -   \\
4DGaussian$^1$  & 28.32  & 0.927 & 28.24  & 0.926 & 23.01 & 0.873  \\
SpacetimeGS$^1$        & 27.42 & 0.915  &  27.45 & 0.920   & - & - \\
LocalDyGS$^2$   & 29.23 & 0.939   & 29.01  &0.931   & 23.21 & 0.875 \\
\ourname(Ours)$^2$        & \textbf{30.61} & 0.948  & \textbf{30.37}  & 0.939    & \textbf{24.78} & \textbf{0.881}  \\
\bottomrule
\end{tabular}
}
\vspace{-1em}
\end{table}

\textbf{MeetRoom \cite{li2022streaming}}. The MeetRoom dataset consists of three forward-facing scenes captured by an array of 11–12 cameras, with a resolution of 1080p and a frame rate of 30 FPS. The dataset is sparse in camera viewpoints, making it well suited for evaluating the performance of our method under sparse-view settings. Following previous work, we use camera 0 for testing.

\textbf{VRU basketball \cite{wuswift4d}}. The VRU dataset is a highly challenging benchmark featuring large-scale motion. Since many existing methods are already overfitted to small-motion scenarios such as N3DV, VRU provides a more suitable testbed for evaluating new approaches. Each VRU scene is captured by 36 cameras and includes three scenes in total: two scenes of 250 frames at 1080p resolution and 25 FPS (VRU DG and GZ), and one scene of 1400 frames at 4K resolution and 25 FPS (VRU Long360). Following prior work, we use camera views \({0, 10, 20, 30}\) for testing, with the remaining views used for training.

\textbf{SelfCap \cite{xu2024longvolcap}}. The SelfCap dataset consists of three dynamic videos, each captured at 60 FPS and 4K resolution using a synchronized array of 22 iPhone cameras. The video lengths range from 2 to 10 minutes, making this dataset particularly well suited for evaluating methods on long video sequences.

\textbf{ PKU-DyMVHumans \cite{zheng2024pkudymvhumansmultiviewvideobenchmark}}. The dataset comprises 45 dynamic human-centric scenarios, totaling approximately 8.2 million frames. It captures 32 professional performers (16 males, 11 females, and 5 children) engaging in four distinct action types: dance, kungfu, sport, and fashion show, across varied locations and clothing styles. The data is categorized into two resolution groups: 36 scenes at 1920×1080 (1080P) with durations of 10 to 487 seconds, and 8 scenes at 3840×2160 (4K Studio) with durations of 10 to 231 seconds.

\begin{table}[t]
    \centering
        \caption{The ablation study on the number of temporal grids $M$ on GZ. }
        \vspace{-0.5em}
    \begin{threeparttable}
    \resizebox{0.6\columnwidth}{!}
    {
        \begin{tabular}{lccccc}
        \toprule
         Metrics &  M=1   & M=4   &  M=8  &  M=12 \\
        \midrule[0.5pt]
        \multirow{1}{*}{PSNR\(\uparrow\)}  & 29.10 & 29.31 & 30.11 & \textbf{30.61} \\
        \multirow{1}{*}{SSIM\(\uparrow\) }  & 0.933 & 0.939 & 0.942 & \textbf{0.948} \\
        \multirow{1}{*}{LPIPS \(\downarrow\)}  & 0.086 & 0.077 & 0.062 & \textbf{0.053} \\
        \bottomrule
        \end{tabular}
    }
\end{threeparttable}
\vspace{-1em}
    \label{tab: abla grids}
\end{table}

\subsection{Evaluation}

\begin{figure*}[h]
    \centering
    \includegraphics[width= \textwidth ,trim=0cm 1cm 0cm 0cm]{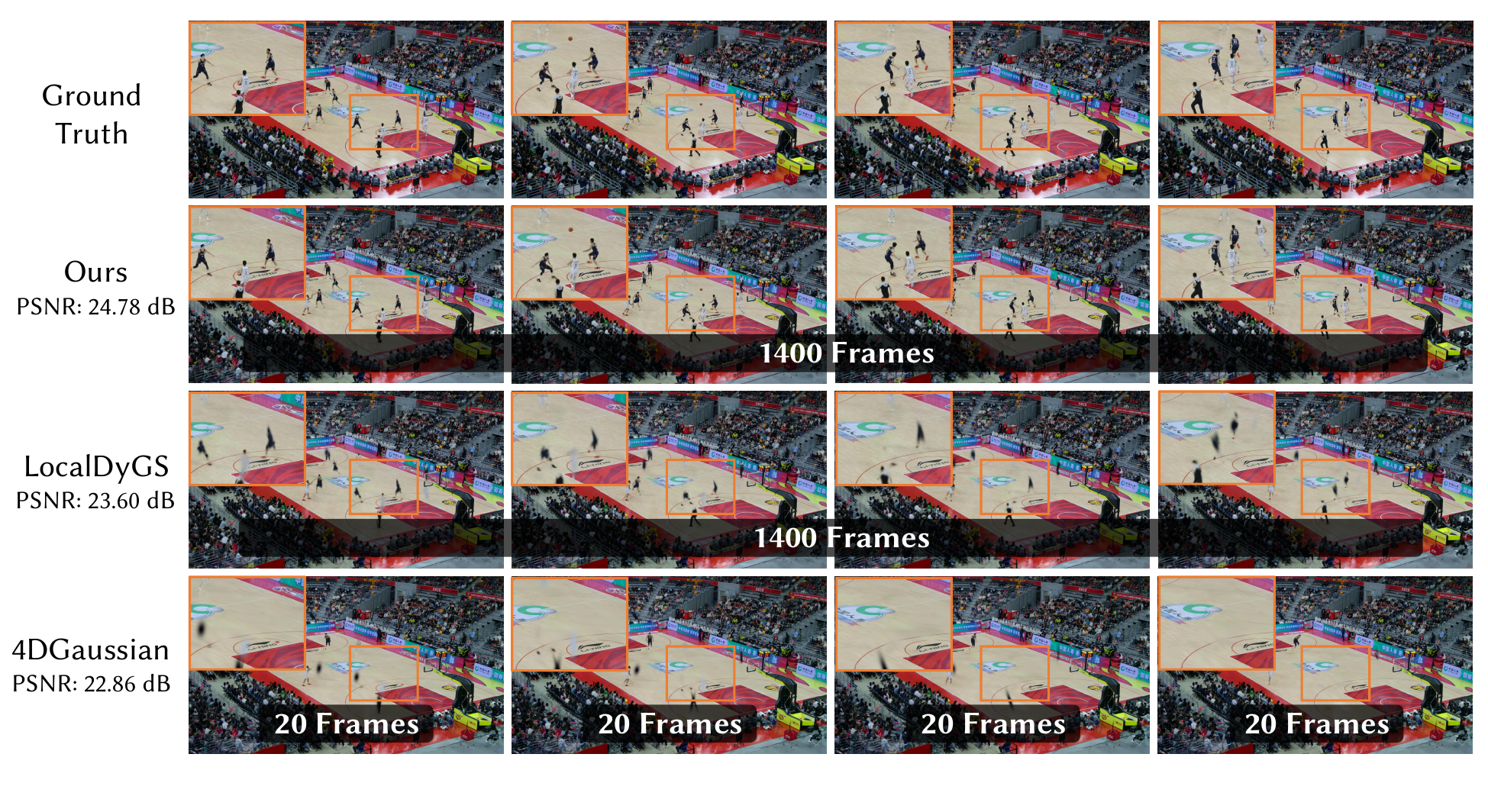}
    \vspace{-2.5em}
    \caption{Qualitative comparisons on the VRU dataset with 1,400 frames (about 1 minute). Our method is capable of training large-scale dynamic scenes with thousands of frames in a single run, while achieving superior rendering quality compared to prior approaches. }
    \label{figs:comparison_vru}
\end{figure*}

\begin{figure*}[t]
    \centering
    \includegraphics[width= 0.9\textwidth ,trim=0cm 1cm 0cm 0cm]{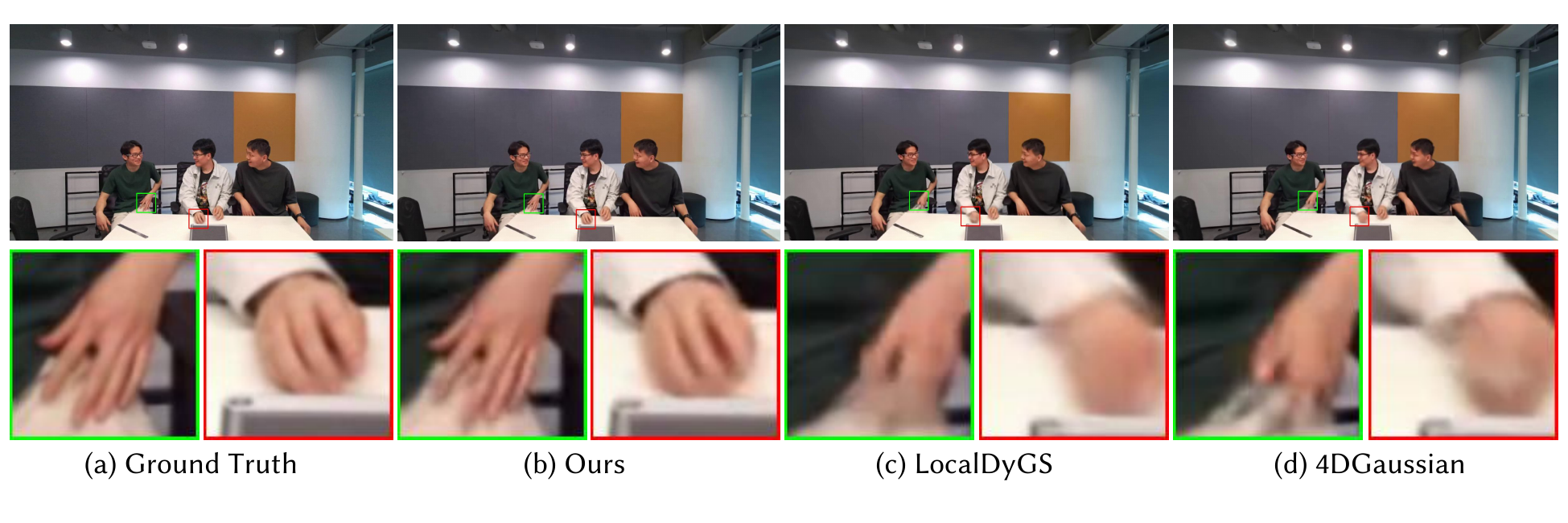}
    \vspace{-0.8em}
    \caption{Qualitative comparisons on the MeetRoom dataset (sparse-view scenario). The results show that our approach preserves more fine details and maintains clarity in regions with large motion compared to prior methods.}
    \label{figs:comparison_meetroom}
    \vspace{-1.2em}
\end{figure*}

We compare our method against a set of representative classical approaches and recent state-of-the-art methods, including 4DGS~\cite{wu20244d}, 3DGStream~\cite{sun20243dgstream}, Grid4D~\cite{xu2024grid4d}, LocalDyGS~\cite{wu2025localdygs}, and Swift4D~\cite{wuswift4d}. These methods span diverse design choices for dynamic scene representation and thus constitute a strong benchmark for evaluation. We conduct quantitative evaluations on three real-world datasets, including the classic N3DV dataset, the VRU dataset with large-scale motion, and the MeetRoom dataset featuring sparse camera views, which together cover a wide range of practical scenarios. Our evaluation focuses on three complementary aspects: image quality measured by PSNR, SSIM, and LPIPS, training and rendering time. SelfVolCap~\citep{xu2024longvolcap} and FreeTimeGS~\citep{wang2025freetimegs} have not released public implementations at the time of writing and are hence not included in our comparisons.

\subsubsection{Quantitative Evaluation}

We first compare our method with prior approaches using standard quantitative metrics across all datasets. As reported in Tab.~\ref{tab: n3dv}, Tab.~\ref{tab: meetroom}, and Tab.~\ref{tab:vru}, our method consistently achieves superior reconstruction quality in terms of PSNR, SSIM, and LPIPS across all datasets, indicating improved fidelity in both pixel accuracy and perceptual quality. Importantly, these gains are achieved without sacrificing efficiency, as our method maintains comparable training time and supports real-time rendering .

\begin{table}[t]
    \centering
        \caption{The ablation study on the of window size $W$ on VRU Long. }
        \vspace{-1em}
    \begin{threeparttable}
    \resizebox{0.7\columnwidth}{!}
    {
        \begin{tabular}{lccccc}
        \toprule
         Metrics &  W=1   &  W=3    &   W=5  &  W=7 &  w/o   \\
        \midrule[0.5pt]
        \multirow{1}{*}{PSNR\(\uparrow\)}  & 24.40 &  \textbf{24.50} &  24.39 &  24.42 &  24.02  \\
        \multirow{1}{*}{DSSIM\(\downarrow\) }  &  0.080 &\textbf{0.079} &  0.080 & 0.080 & 0.085 \\
        \multirow{1}{*}{LPIPS \(\downarrow\)}  & 0.147 & \textbf{0.145} &  0.147 & 0.147 & 0.153 \\
        \bottomrule
        \end{tabular}
    }
    \vspace{-3em}
\end{threeparttable}
    \label{tab: abla window}
\end{table}

Beyond reconstruction quality, \textbf{our method demonstrates a clear advantage in modeling long temporal sequences}. On the N3DV benchmark, existing methods are typically limited to short clips of up to 300 frames, whereas our approach enables one-shot reconstruction of sequences with up to 1200 frames. Similarly, on the VRU dataset, our method successfully models sequences of up to 1400 frames, while prior approaches \cite{wu2025localdygs} usually handle only around 20 frames, corresponding to an approximately 70$\times$ increase in temporal coverage. These results highlight the scalability of our approach to long and complex dynamic scenes. On SelfCap\cite{xu2024longvolcap} dataset, the results can be seen in Tab.\ref{tab:selfcap_results}.

\subsubsection{Qualitative Evaluation}

Qualitative comparisons further validate the advantages of our method across different scenarios. As shown in Fig.~\ref{figs:comparison_n3dv} and Fig.~\ref{figs:comparison_meetroom}, our reconstructions preserve fine-grained details with higher visual fidelity, particularly in challenging regions such as human hands, where competing methods often exhibit noticeable blurring or loss of structure. For scenes with complex and fast motion, such as the rapidly moving athletes in Fig.~\ref{figs:comparison_vru}, our method is able to reconstruct dynamic content with high fidelity over long temporal spans. In contrast, LocalDyGS~\citep{wu2025localdygs} suffers from missing body parts, while streaming-based approaches such as 4DGS~\citep{wu20244d} exhibit more severe degradation, including increased blur and missing geometry, due to accumulated errors over time. Notably, our results are obtained by reconstructing the entire long sequence in a single training process, which is particularly challenging for long and complex dynamic scenes.

\subsubsection{Generalization experiment}

We conducted comprehensive experiments on large-scale datasets, including PKU-DyMVHumans and SelfCap~\citep{xu2024longvolcap}. These datasets contain challenging scenarios with complex human motions and long video sequences. The extensive evaluation demonstrates the strong generalization capability of our method in dynamic scene modeling. Additional results (such as SelfCap~\citep{xu2024longvolcap}, PKU-DyHuman \cite{zheng2024pkudymvhumansmultiviewvideobenchmark}) are provided in the \textbf{supplementary videos}.

\begin{table}[t]
    \centering
        \caption{The ablation study on the number of keyframes $K$(conducted on VRU Long 250 frames). }
         \vspace{-1em}
    \begin{threeparttable}
    \resizebox{0.6\columnwidth}{!}
    {
        \begin{tabular}{lccccc}
        \toprule
         Metrics &  K=250   & K=125   &  K=25  &  K=12 \\
        \midrule[0.5pt]
        \multirow{1}{*}{PSNR\(\uparrow\)}  & \textbf{24.42} & 24.30 & 24.17 & 23.76 \\
        \multirow{1}{*}{DSSIM\(\downarrow\) }  & \textbf{0.080} & 0.081 & 0.083 & 0.086 \\
        \multirow{1}{*}{LPIPS \(\downarrow\)}  & \textbf{0.147} & 0.150 & 0.157 & 0.165 \\
        \bottomrule
        \end{tabular}
    }
    \vspace{-1em}
\end{threeparttable}
    \label{tab: abla k frames}
\end{table}

\begin{table}[t]
    \centering
        \caption{The ablation study on different features on VRU Long. }
        \vspace{-1em}
    \begin{threeparttable}
    \resizebox{0.7\columnwidth}{!}
    {
        \begin{tabular}{lccccc}
        \toprule
         Metrics &  PSNR $\uparrow$   & DSSIM1 $\downarrow$   &  DSSIM2 $\downarrow$ &  LPIPS $\downarrow$ \\
        \midrule[0.5pt]
        \multirow{1}{*}{$f_a+f_t$}  & 24.30 & 0.082 & 0.038 & 0.151 \\
        \multirow{1}{*}{$f_s+f_t$}  & 22.58 & 0.116 & 0.055 & 0.240 \\
        \multirow{1}{*}{$f_a+f_s+f_t$}  & \textbf{24.42} & \textbf{0.080} & \textbf{0.037} & \textbf{0.147} \\
        \bottomrule
        \end{tabular}
    }
\end{threeparttable}
    \vspace{-1.5em}
    \label{tab: abla features}
\end{table}

\begin{table}[t]
    \centering
    \caption{ Quantitative results on the SelfCap dataset (Bar) (test views: 1, 7, 13; 2000 frames).}
    \vspace{-1em}
    \resizebox{0.7\columnwidth}{!}{
    \begin{tabular}{lccc}
        \toprule
        Method & PSNR$\uparrow$ & DSSIM$\downarrow$ & LPIPS$\downarrow$ \\
        \midrule
        4DGaussian & 27.86 & 0.0772 & 0.145 \\
        Ours       & \textbf{29.13} & \textbf{0.0689} & \textbf{0.110} \\
        \bottomrule
    \end{tabular}
    }
    \label{tab:selfcap_results}
\end{table}

\begin{table}[t]
    \centering
    \caption{ Inference time breakdown on VRU-GZ.}
    \vspace{-1em}
    \begin{threeparttable}
    \resizebox{0.9\columnwidth}{!}{
        \begin{tabular}{lccccc}
        \toprule
        & Feature Extraction & MLP Decoding & Rendering & Others & Full \\
        \midrule
        Time (ms) & 0.7 & 5.9 & 4.4 & 4.6 & 15.6 \\
        \bottomrule
        \end{tabular}
    }
    
    \end{threeparttable}
    \vspace{-1.5em}
    \label{tab:details runtime}
\end{table}

\subsection{Ablation Study}


\textbf{The Number of Temporal Grids.}
We evaluate the impact of the number of temporal grids \(M\) on the 250-frame VRU-gz sequence by setting \(M={1,4,8,12}\). As shown in Tab.~\ref{tab: abla grids}, the rendering quality consistently improves with larger \(M\), indicating a positive correlation between performance and temporal grid capacity. Increasing the number of grids enables the model to store richer temporal information, resulting in more accurate and stable reconstructions, which aligns with our design motivation. 

\textbf{The ablation of query window size \(W\).}
We further analyze the effect of query window size \(W\) on the VRU-long dataset. As reported in Tab.~\ref{tab: abla window}, the performance remains relatively stable across different window sizes, suggesting that our method is robust to this hyperparameter and does not heavily rely on a specific temporal context range.  However, removing the temporal window leads to a noticeable degradation in rendering quality, which validates the importance of the proposed query window.

\textbf{The number of keyframes.} We investigate the influence of the number of keyframes on reconstruction quality. By varying the number of selected keyframes while keeping all other settings fixed (in Tab. \ref{tab: abla k frames}), we observe that increasing the number of keyframes consistently improves performance, as it provides stronger temporal coverage and more reliable supervision for long sequences.

\textbf{Ablation of static feature \( f_s \) and anchor feature \( f_a \).}
To assess the effectiveness of the proposed static feature \( f_s \) and anchor feature \( f_a \), we perform an ablation study in Tab.~\ref{tab: abla features} by removing each component while keeping all others unchanged. Excluding either \( f_s\) or \( f_a \) leads to degraded reconstruction quality, particularly in regions with large motion and fine geometric details. These results indicate that both features are critical for encoding spatial context and stabilizing feature aggregation across frames.

\textbf{Inference time details.} To further analyze the time consumption of our model, we report a detailed breakdown in Tab.~\ref{tab:details runtime}.

\begin{figure*}[t]
    \centering
    \includegraphics[width= \textwidth ,trim=0cm 1cm 0cm 0cm]{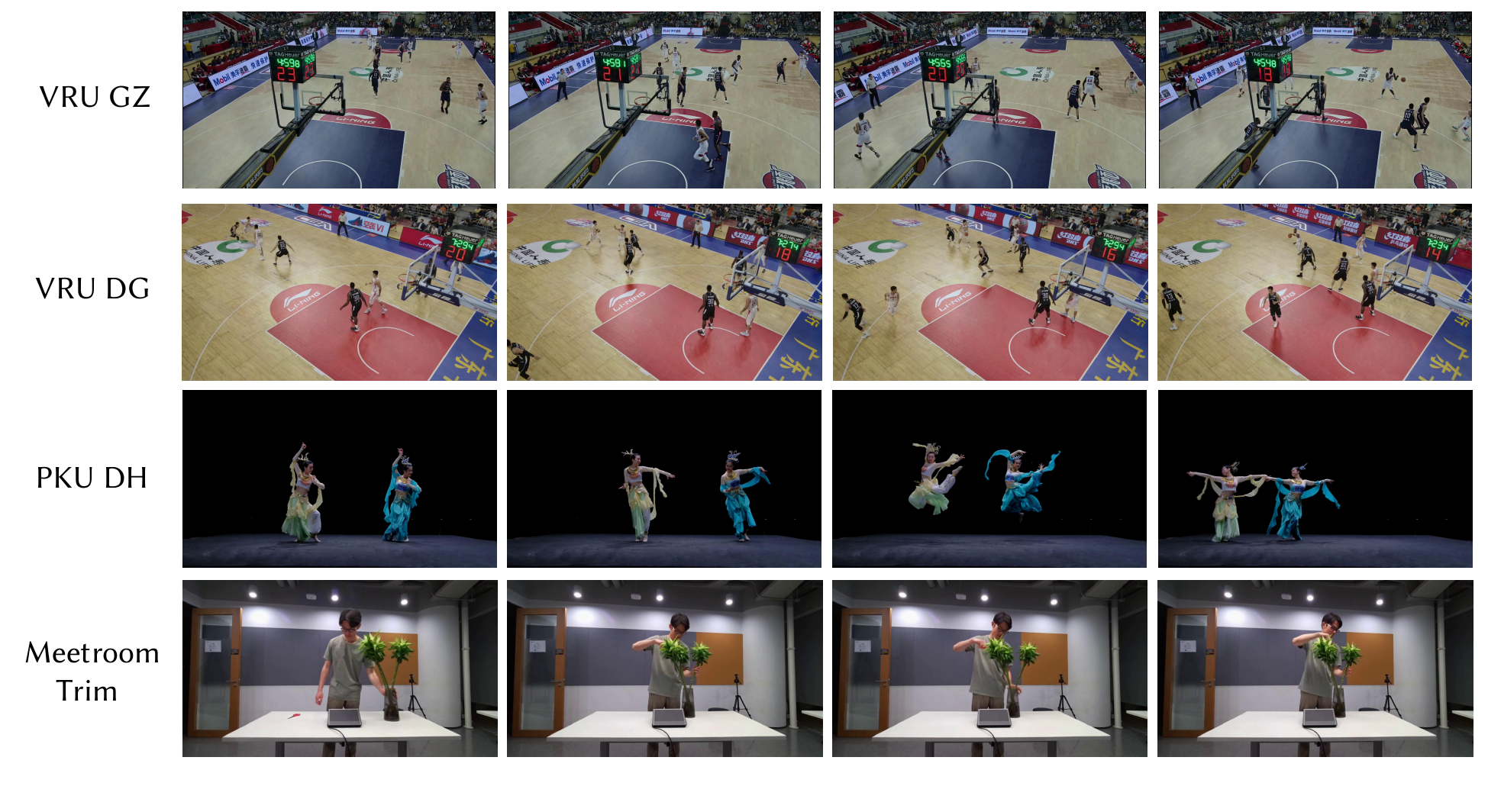}

    \caption{Qualitative results on a broader set of datasets.}
    \label{figs:More}
\end{figure*}

\section{Discussion and Limitations}

Similar to previous works \cite{xu2024longvolcap,wu2025localdygs}, our method follows an offline reconstruction paradigm and therefore does not support real-time processing, which remains an open and challenging problem in volumetric video reconstruction. In addition, our approach relies on camera poses and sparse point clouds estimated by COLMAP. Although COLMAP provides accurate and robust geometric initialization, its relatively high computational cost has become a common bottleneck for many recent 3DGS-based pipelines. With the rapid progress of feed-forward camera pose estimation and scene reconstruction methods, we believe this limitation will be gradually mitigated in future systems.

Furthermore, we observe mild temporal jitter in regions associated with extremely sparse viewpoints, such as distant audience areas in the VRU dataset. This behavior is mainly caused by insufficient multi-view observations in these regions, which leads to \textbf{under-constrained optimization during training} and makes it difficult to consistently enforce temporal coherence. Similar phenomena have also been reported in static reconstruction methods, where Gaussians corresponding to marginal viewpoints are often blurry due to limited observational constraints. Notably, these artifacts are restricted to peripheral areas that are weakly observed and lie outside the primary focus of our reconstruction and evaluation. Consequently, they have limited impact on the reconstruction quality of the main subjects or the overall conclusions of this work. We leave the incorporation of stronger temporal regularization or additional geometric priors for sparsely observed regions as an interesting direction for future research.

Another key factor affecting dynamic reconstruction quality is the limited multi-view coverage in dynamic datasets. These datasets typically provide only 10–40 views. For example, in the VRU dataset, cameras are mainly distributed around the stadium boundary in a large-scale scene, resulting in weak geometric constraints in the inner field region. In addition, dynamic subjects occupy only a small number of pixels, making gradient-based optimization particularly challenging.

\vspace{-0.8em}
\section{Conclusion}
We propose \ourname, a novel framework capable of volumetric video reconstruction for long multi-view sequences with large-scale motion, extending reconstruction from small-motion, short sequences to long sequences with significant motion. Our approach is built on two core ideas: time-conditioned anchors and feature decomposition. Time-conditioned anchors localize all moving objects in space-time and specify their corresponding timestamps. To map anchors to Gaussians, we learn a set of features for each 
anchor, including anchor features for fine-detail control, static features for spatial stability, and temporal features for temporal modeling. By fusing these features, our model can stably reconstruct volumetric videos spanning thousands of frames while achieving state-of-the-art performance in both objective and perceptual quality.

\section{Acknowledges}
This work is financially supported by Fundamental and Interdisciplinary Disciplines Breakthrough Plan of the Ministry of Education of China(Grant No. JYB2025XDXM413), Guangdong Provincial Key Laboratory of Ultra High Definition Immersive Media Technology(Grant No. 2024B1212010006), this work is also financially supported for Outstanding Talents Training Fund in Shenzhen, Shenzhen Science and Technology Program(Grant No. SYSPG20241211173 440004  and KJZD20230923114707016), R24115SG MIGU-PKU META VISION TECHNOLOGY INNOVATION LAB.

\clearpage
\balance
\bibliographystyle{ACM-Reference-Format}
\bibliography{main}



\end{document}